\documentclass[lettersize,journal]{IEEEtran}
\usepackage{amsmath,amsfonts}
\usepackage{algorithmic}
\usepackage{algorithm}
\usepackage{array}
\usepackage[caption=false,font=normalsize,labelfont=sf,textfont=sf]{subfig}
\usepackage{textcomp}
\usepackage{stfloats}
\usepackage{url}
\usepackage{verbatim}
\usepackage{graphicx}
\usepackage{cite}
\usepackage{booktabs} 
\usepackage{pifont}
\begin{document}

\title{PointMT: Efficient Point Cloud Analysis with Hybrid MLP-Transformer Architecture}

\author{Qiang Zheng, Chao Zhang, Jian Sun*
}

\markboth{Journal of \LaTeX\ Class Files,~Vol.~14, No.~8, August~2021}%
{Shell \MakeLowercase{\textit{et al.}}: A Sample Article Using IEEEtran.cls for IEEE Journals}


\maketitle

\begin{abstract}
In recent years, point cloud analysis methods based on the Transformer architecture have made significant progress, particularly in the context of multimedia applications such as 3D modeling, virtual reality, and autonomous systems. However, the high computational resource demands of the Transformer architecture hinder its scalability, real-time processing capabilities, and deployment on mobile devices and other platforms with limited computational resources. This limitation remains a significant obstacle to its practical application in scenarios requiring on-device intelligence and multimedia processing. To address this challenge, we propose an efficient point cloud analysis architecture, \textbf{Point} \textbf{M}LP-\textbf{T}ransformer (PointMT). This study tackles the quadratic complexity of the self-attention mechanism by introducing a linear complexity local attention mechanism for effective feature aggregation. Additionally, to counter the Transformer's focus on token differences while neglecting channel differences, we introduce a parameter-free channel temperature adaptation mechanism that adaptively adjusts the attention weight distribution in each channel, enhancing the precision of feature aggregation. To improve the Transformer's slow convergence speed due to the limited scale of point cloud datasets, we propose an MLP-Transformer hybrid module, which significantly enhances the model's convergence speed. Furthermore, to boost the feature representation capability of point tokens, we refine the classification head, enabling point tokens to directly participate in prediction. Experimental results on multiple evaluation benchmarks demonstrate that PointMT achieves performance comparable to state-of-the-art methods while maintaining an optimal balance between performance and accuracy. This research provides an innovative solution for efficient point cloud analysis, offering significant potential for multimedia applications and other domains.

\end{abstract}



\begin{IEEEkeywords}
point cloud, classification, segmentation, MLP, Transformer.


\end{IEEEkeywords}


\section{Introduction}
\IEEEPARstart{W}ith the increasing prevalence of point clouds in real-world applications such as autonomous driving, robotics, and multimedia applications including 3D modeling, virtual reality, and augmented reality, deep learning-based point cloud analysis has gained significant popularity. However, due to the unorderedness, irregularity, and sparsity of point clouds, directly extending successful deep networks from 2-dimensional (2D) image processing to 3-dimensional (3D) data is a non-trivial task. These challenges necessitate innovative approaches to ensure efficient and effective point cloud analysis suitable for multimedia processing and other demanding applications.

Early approaches attempted to transform point clouds into projections~\cite{2015Multi, multi2021} or voxels~\cite{VoxNet2015, SPLATNet2018, OctNet2017}, enabling processing by standard CNNs. However, these methods suffer from geometric information loss and increased storage and computational demands. Recently, models have emerged that directly operate on raw point cloud data. These can be broadly categorized based on the feature extraction operators: Multi-layer Perceptrons (MLPs) based~\cite{PointNetplus2017}, Convolutional Neural Networks (CNNs) based~\cite{PointConv2019}, Graph Neural Networks (GNNs) based~\cite{DGCNN2019}, and Transformer-based~\cite{PT2021}.

Although the Transformer architecture, with its self-attention mechanism, has demonstrated significant advantages in point cloud analysis, it also has inherent limitations. (i) Firstly, the computational complexity of the Transformer is relatively high. For example, the ViT~\cite{2021ViT} employs a global attention mechanism that results in quadratic computational complexity with respect to the number of tokens. Subsequent models, such as the Swin Transformer~\cite{2021Swin}, have mitigated this issue by using local attention mechanisms, but the complexity within local image patches remains quadratic. (ii) Secondly, while the Transformer model emphasizes spatial attention to distinguish the importance of tokens, it often overlooks differences between channels. A dual attention mechanism~\cite{2022DaViT, 2023DTNet} can address both spatial and channel attention, but this approach requires additional attention modules for channels. While this can enhance performance, it also significantly increases computational demands. (iii) Moreover, the field of point cloud research lacks large datasets like ImageNet~\cite{ImageNet2009}. The Transformer model relies on a large number of parameters and substantial data to learn effective feature representations. Consequently, directly applying the Transformer to point cloud analysis may result in slow convergence and limited performance improvement.

This paper proposes a novel method based on the Transformer framework, centered around the MLP-Transformer (MT) Block. This block introduces a local attention mechanism with linear complexity, transforming the traditional self-attention mechanism into an attention mechanism between the central point and its neighboring points. In this mechanism, the central point's token acts as the Query (Q), while the neighboring points' tokens serve as Key (K) and Value (V). Assuming the point cloud contains $N$ points and the number of neighboring points is $k$, the proposed local attention mechanism reduces the computational complexity from $\mathcal{O}(Nk^2)$ to $\mathcal{O}(Nk)$, achieving linear complexity in attention computation and significantly enhancing the network's operational efficiency and scalability.

To address the traditional Transformer's tendency to overlook differences between channels, the MT-Block introduces a parameter-free temperature adaptation mechanism. When the attention scores are same across all channels, this mechanism automatically adjusts the temperature parameter of softmax based on the activation strength and feature diversity within the channels, flexibly adjusting the weight distribution for each channel.

Furthermore, to counter the impact of the lack of large point cloud datasets on the Transformer's convergence speed and performance, the MT-Block adopts a hybrid mechanism combining MLP and Transformer. The MLP model is characterized by rapid convergence and high computational efficiency in point cloud analysis but lacks fine-grained feature perception; in contrast, the Transformer model excels at capturing fine-grained features but converges more slowly. The combination of MLP and Transformer leverages their complementary advantages in feature representation, accelerating convergence speed and reducing computational consumption.

In addition to the MT-Block, this paper designs a new classification head for classification tasks. Within the hierarchical structure, point features also incorporate global information through an increasingly expanded receptive field. The proposed classification head takes shape features and point tokens as inputs and fuses their outputs. Compared with traditional classification heads, this design enables point features to directly participate in shape prediction, enhancing the perception of global shape and thereby improving training efficiency.

In conclusion, this paper makes the following key contributions:
\begin{itemize}
\item We propose a local attention mechanism with linear complexity, significantly enhancing the computational efficiency of the Transformer model.
\item We introduce a channel temperature adaptation mechanism that dynamically adjusts the softmax temperature parameter based on the activation intensity and feature diversity of each channel, thereby enhancing the representational capacity of features.
\item We designed the MT-Block, which integrates the strengths of MLP and Transformer, effectively accelerating model training convergence and enhancing stability.
\item We developed a novel classification head for classification tasks that allows point features to directly participate in the classification process, improving the perception of global shape.
\item Based on the MT-Block, we constructed an efficient point cloud analysis architecture, PointMT, which matches the performance of state-of-the-art models while significantly reducing complexity.
\end{itemize}

\section{Related works}	
\subsection{Transformers in Computer Vision}
The Transformer architecture has revolutionized computer vision (CV), demonstrating remarkable effectiveness across various tasks. \textbf{Foundations and Advancements:} The foundational work~\cite{2017Attention} introduced the Transformer, paving the way for its application in CV. Dosovitskiy et al.~\cite{2021ViT} established the Vision Transformer (ViT), demonstrating its success on image patches for tasks like classification. Subsequent advancements explored improved efficiency, contextual modeling, and pre-training strategies for Transformers in CV. For instance, Swin Transformer~\cite{2021Swin} utilized shifted windows to enhance efficiency and capture long-range dependencies, while BEiT~\cite{2022BEiT} leveraged BERT-style pre-training for powerful self-supervised learning capabilities. \textbf{Object Detection:} DETR~\cite{2020DETR} pioneered the use of Transformers for object detection, eliminating the need for hand-crafted components like non-maximal suppression (NMS) and anchor boxes. Deformable DETR~\cite{2021DeformDETR} and Conditional DETR~\cite{2021CondDETR} further enhanced DETR's performance through deformable attention mechanisms and conditional spatial queries, respectively. Sparse DETR~\cite{2022SparDETR} focused on improving computational efficiency by utilizing sparse Transformers. \textbf{Segmentation:} SETR~\cite{2021SETR} was one of the first Transformer-based approaches for semantic segmentation, followed by advancements like RPM~\cite{2022RPM} demonstrating success in weakly supervised settings. PVT~\cite{2021PVT} introduced a hierarchical architecture that effectively aggregates features across different scales, achieving competitive results on various segmentation benchmarks. \textbf{Video Understanding:} Literature~\cite{2022TMVLU} provides a comprehensive review of Transformer's performance in image and video tasks, highlighting its attention mechanism and global modeling ability. TokShift~\cite{2021TokShift} introduces the Token Shift Module, a computational efficient operator for modeling temporal relations in video signals, which significantly improves the performance of the Transformer in video classification tasks. The review article~\cite{2023UVTS} surveys the use of Transformer in video segmentation, discussing its interpretability and the state-of-the-art models. LSTCL~\cite{2022LSTCL} presents a self-supervised pretraining method, Long-Short Temporal Contrastive Learning, which enhances the performance of video Transformers in action recognition tasks.

\subsection{Point Cloud Analysis with Deep Learning}
Recent research in point cloud analysis has seen significant advancements in deep learning approaches, particularly with the exploration of different network architectures like MLP, CNN, GNN, and Transformer.

\textbf{MLP-based methods} have played a crucial role in establishing the foundation for point cloud learning. Pioneering works like PointNet~\cite{PointNet2017} and PointNet++~\cite{PointNetplus2017} introduced efficient ways to learn from unordered point sets using shared MLPs. Recent contributions such as PointContrast~\cite{2020PointContrast}, PointAugment~\cite{2020PointAugment}, and PointClustering~\cite{2023PointClustering} delve deeper into pre-training strategies and data augmentation techniques to improve performance and robustness.

\textbf{CNN-based methods} offer an alternative approach by leveraging convolutional operations on local patches. Recent advancements in point cloud CNNs include PointConv~\cite{2018PointCNN} with its learnable $\chi$-transformation for feature weighting and potential reordering. In contrast, SpiderCNN~\cite{2018SpiderCNN} constructs kernels by combining learned functions. To address the challenge of non-grid data, PCCN~\cite{2021Deep} tackles non-grid data with a learnable, continuous space operator. KPConv~\cite{2019KPConv} utilizes learnable point sets in Euclidean space, where each input point determines its filter via linear combination. Alternatively, PAConv~\cite{2021PAConv} leverages a pre-defined Weight Bank and trainable ScoreNet for weight coefficient learning. A-CNN~\cite{2019ACNN} defines kernels based on input points within annular regions for efficient local geometric feature capture.

\textbf{GNN-based methods} model the inherent relationships between points in a cloud, making them particularly powerful. DGCNN~\cite{DGCNN2019} leveraged edge convolutions within a dynamic graph, enabling the capture of evolving relationships between points. Point2Node~\cite{Point2Node2019} introduced a gating mechanism that dynamically integrates node relationships for adaptive feature aggregation at the channel level. The focus on adaptability continues with AdaptConv~\cite{Adaptive2021}, which proposes a method for generating convolution kernels that adapt to point features, allowing for the capture of diverse relationships within the point cloud. Recent advancements further highlight the effectiveness of hierarchical structures and message passing mechanisms in GCNs for point clouds. Works like HPGNN ~\cite{2022HPGNN}, CoFiNet~\cite{2021CoFiNet}, and Point-GNN~\cite{2020Point-GNN} demonstrate the promise of these approaches for tasks such as registration, object detection, and classification.

\textbf{Transformer-based methods} have recently gained significant traction. Pioneering work PCT~\cite{PCT2021} laid the groundwork for this exciting new direction. These methods leverage self-attention mechanisms, a core principle of Transformers, to capture complex relationships within point clouds. However, applying global self-attention directly to point clouds presents a challenge – the computational cost scales quadratically with the number of points, leading to significant memory and processing bottlenecks. To overcome this limitation, Point Transformer~\cite{point-trans2021} introduced a local attention mechanism inspired by SAN~\cite{SAN2020}. This approach focuses on capturing local geometric features, significantly reducing computational complexity. Building upon this foundation, Point Transformer V2~\cite{point-transV2-2022} further refines the local attention strategy. It introduces group vector attention and position encoding multipliers to more effectively aggregate features from neighboring points. Additionally, partition-based pooling is proposed to enhance spatial alignment and sampling efficiency. Stratified Transformer~\cite{Stratified2022} also utilizes a local attention mechanism, but with a twist. It employs a two-stage sampling approach: dense sampling for nearby points and sparse sampling for distant points. This strategy allows the model to expand its receptive field while maintaining computational efficiency, demonstrating exceptional performance in point cloud segmentation tasks. CDFormer~\cite{CDFormer2023} begins by capturing short-range dependencies through local self-attention. It then samples a set of reference points to learn global, long-range dependencies. Finally, these dependencies are propagated to each point, enriching the contextual information of each token within the point cloud. DTNet~\cite{2023DTNet} introduces a dual attention mechanism for point cloud processing, effectively capturing spatial and channel attention through its Dual Point Cloud Transformer (DPCT) module, thereby enhancing the semantic understanding of contextual dependencies.

\section{Methodology}
In this chapter, Sec. \ref{sec-overview-arch} presents an overview of the PointMT architecture, which is constructed upon the MLP-Transformer hybrid framework. Sec. \ref{sec-linear-attn} delves into the rationale for selecting local attention for this study, outlining measures to address its limitation in capturing long-range dependencies. Additionally, it provides detailed insights into the design of a local attention mechanism with linear complexity, which constitutes a key contribution of this work. Sec. \ref{sec-temp-adapt} addresses the issue that traditional attention mechanisms overlook the differences between channels, we propose a parameter-free temperature adaptation mechanism. This mechanism can flexibly adjust the softmax temperature parameters for each channel, optimizing the weight distribution and enhancing the model's representational capacity for channel features. \ref{sec-MT-hybrid} introduces an concise MLP-Transformer hybrid architecture. This architecture aims to improve convergence rates and enrich the contextual information of local features, thus mitigating the challenges of slow convergence and insufficient training of Transformer models. Sec. \ref{sec-SPF-Head} demonstrates a classification head that integrates point and shape features, allowing point features to directly contribute to classification prediction. This facilitates feature alignment at both the point and shape levels, leading to performance improvements.

\subsection{Overview of Network Architecture} \label{sec-overview-arch}
\begin{figure}[htbp]
    \centering
    \includegraphics[width=\linewidth]{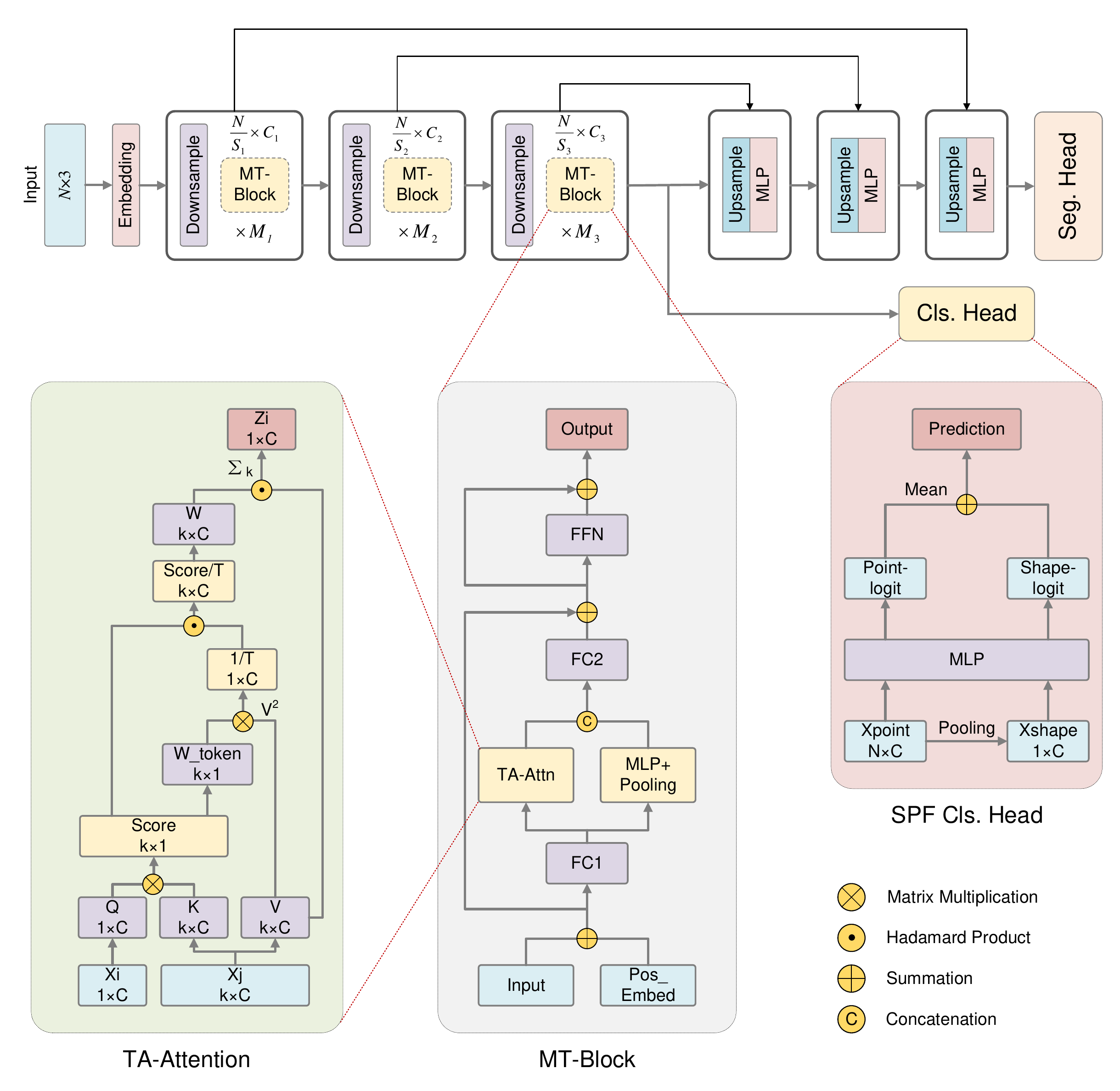}
    \caption{An overview of the PointMT network architecture and its key components. The diagram illustrates the TA-Attention module, which integrates a linear local attention mechanism (see Sec. \ref{sec-linear-attn}) and a channel \textbf{temperature} \textbf{adaptation} strategy (see Sec. \ref{sec-temp-adapt}) to enhance feature representation. Additionally, it features the MT-Block, an innovative hybrid unit that combines \textbf{MLP} and \textbf{Transformer} architectures (see Sec. \ref{sec-MT-hybrid}), as well as the SPF Cls. Head (see Sec. \ref{sec-SPF-Head}), a novel classification head designed based on the \textbf{shape-point-fusion} mechanism to improve classification accuracy.}
    \label{fig-Component-Net}
\end{figure}

This section provides an overview of PointMT, designed for our proposed classification and semantic segmentation models, as depicted in Fig. \ref{fig-Component-Net}. The key components of PointMT are elaborated in the following subsections.

\subsubsection{Classification Model}
The classification model comprises an encoder and a classifier. The encoder adopts a hierarchical pyramid structure, with each stage housing a downsampling layer and multiple MT-Blocks. The resulting features from the encoder are directed to the classification head (the new classification head elaborated in Sec. \ref{sec-SPF-Head}) to yield predictions.

\subsubsection{Semantic Segmentation Model}
The segmentation model employs a symmetric encoder-decoder architecture. While the encoder structure mirrors that of the classification model, it includes a larger number of stages and MT-Blocks to accommodate the larger dataset and the need for extracting more detailed abstract semantic information in the semantic segmentation task. In the decoder of the segmentation model, multiple stages are present, each comprising an upsampling layer and an MLP module. These upsampled features are combined with the output features from the corresponding encoder stage through skip connections before being processed in the MLP module. Afterwards, the decoder's output is passed to the segmentation head for predicting semantic labels.

\subsection{Linear Complexity Local Attention Mechanism}\label{sec-linear-attn}

\begin{figure}[ht]
    \centering
    \includegraphics[width=0.5\linewidth]{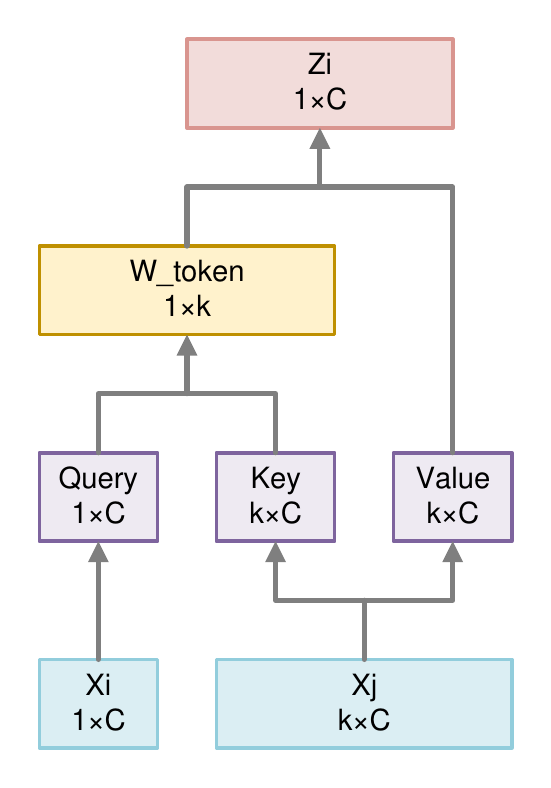}
    \caption{Local attention with linear complexity proportional to the number of points $N$ and the size of local neighborhoods $k$.}
    \label{fig-LinearAttn}
\end{figure}

Based on the local attention mechanism, this study updates the central token using adjacent tokens, thereby developing an attention mechanism with linear complexity. Fig. \ref{fig-LinearAttn} illustrates the computation process of linear local attention. 

In Fig. \ref{fig-LinearAttn}, assuming $X = (P, F)$ constitutes a complete point cloud scene, where each point is denoted as $x_i = (p_i, f_i) \in X$, with $p_i \in \mathbb{R}^3$ representing point coordinates and $f_i \in \mathbb{R}^C$ denoting point features. For local feature extraction, a neighborhood $\mathcal{N}(x_i) = x_j = {(p_j, f_j) | p_j \in N(p_i)}$ is obtained for a specific point $x_i$. The local attention proposed in this paper aims to update the central point's features utilizing geometric information from its neighborhood. Thus, the local attention herein designates the central point as the Query point and the collection of neighborhood points as the Key, effectively aggregating neighborhood information while substantially reducing computational load. Linear transformations are applied to $x_i$ and $x_j$ to yield $Q = \text{Linear}_Q(f_i)$, $K = \text{Linear}_K(f_j - f_i)$, and $V = \text{Linear}_V(f_j - f_i)$, where $f_i \in \mathbb{R}^{1 \times C}$ and $f_j \in \mathbb{R}^{k \times C}$, with $k$ representing the number of neighborhood points. Subsequently, attention weights are computed, and neighborhood features are aggregated as depicted below:
\begin{equation}
\label{eq-score-token}
\text{Score}(x_i) = {QK^T} / {\sqrt{C}}
\end{equation}

\begin{equation}
\label{eq-weight-token}
W_{\text{token}} = \text{Softmax}(\text{Score}(x_i))
\end{equation}

\begin{equation}
\label{eq-attn-token-output}
Z = W_{\text{token}}V
\end{equation}

In Eq. \ref{eq-score-token} to \ref{eq-attn-token-output}, the symbol ``$\text{Score}$'' denotes the attention score, and the subscript ``$\text{token}$'' signifies spatial attention. The symbol ``$Z$'' denotes the output neighborhood aggregation feature. This operation of aggregating information from $k$ neighborhood points to a single central point formally aligns with pooling methods for acquiring local features. 

Comparing the proposed attention mechanism with the traditional self-attention mechanism, an analysis of their computational complexity is presented. Assuming $N$ points in the point cloud scene, the computational complexity of global self-attention is $\mathcal{O}(N^2)$. For local self-attention, assuming a neighborhood size of $k$, the overall computational complexity is $\mathcal{O}(Nk^2)$. In contrast, the self-attention mechanism proposed herein reduces the $k$ Query point to a single central point, thereby reducing the complexity to $\mathcal{O}(Nk)$, establishing a linear relationship with both the scale of the point cloud and the neighborhood size.

\subsection{Temperature Adaption Attention Mechanism} \label{sec-temp-adapt}

The conventional dual attention mechanism, while enhancing feature processing capabilities, inevitably leads to a significant increase in computational expense. Traditional dual attention is commonly implemented using specialized modules equipped with learnable parameters for channel attention. Although this method explicitly targets informative channels, it comes with certain drawbacks. The traditional dual attention mechanism~\cite{2022DaViT, 2023DTNet} amplifies the model's spatial and time complexity by introducing additional parameters, making it more arduous to train and deploy, especially in resource-constrained settings. Furthermore, the proliferation of parameters heightens the risk of model overfitting. 

This paper introduces a novel parameter-free Temperature Adaption Attention (TA-Attention) mechanism that directly modulates spatial attention scores, thereby optimizing the distribution of attention weights across channels (refer to Fig. \ref{fig-Component-Net}). On the basis of obtaining $\text{Score}(x_i)$ and $W_{\text{token}}$ from Eq. \ref{eq-score-token} and Eq. \ref{eq-weight-token}, the TA-Attn mechanism performs the following calculations:

\begin{equation}
\overline{V^2} = W_{\text{token}} V^2
\end{equation}

\begin{equation}
T = \frac {1} {\overline{V^2} / \sqrt{k}}
\end{equation}

\begin{equation}
W = \text{Softmax}(\text{Score}(x_i)/T)
\end{equation}

\begin{equation}
Z = \sum_{i=1}^{k} (W \odot V)
\end{equation}

In the equations above, $W_{\text{token}}$, $V$, and $k$ are consistent with their definitions in  Eq. \ref{eq-score-token} and Eq. \ref{eq-weight-token}. Considering $W_{\text{token}}$ as a set of probability distributions, the length of $\overline{V^2}$ matches the number of channels $C$, representing the weighted average of $V^2$ within each channel. After scaling $\overline{V^2}$ by dividing it by $\sqrt{k}$, its reciprocal is taken as the temperature parameter $T$ for the softmax function. Although the initial score is the same for all channels, the differing parameters $T$ for each channel result in distinct weight distributions. Computationally, TA-Attention evolves from the linear attention mechanism introduced in Sec. \ref{sec-linear-attn}, by interposing an operation to calculate the temperature parameter and adjust the attention scores between Eq. \ref{eq-score-token} and Eq. \ref{eq-weight-token}.

The effectiveness of TA-Attention can be qualitatively explained by the impact of the temperature parameter on the softmax distribution. The TA-Attention model introduces a temperature parameter $T$ to modulate the output distribution of the softmax function, thereby achieving dynamic adjustment of feature selectivity. 
Suppose there are $k$ tokens in a local region, and the channel length is $C$. The weights of spatial attention is denoted as $W_{\text{token}} = \{ w_i \}_{i \in [0, k]}$, with the same weight distribution across all channels. Then, for a specific channel $j$, the weighted average value of $V$ is given by

\begin{equation}
\overline{v_{j}} = \sum_{i=1}^{k} w_i v_{ij},
\end{equation}
and the feature diversity within the channel is represented as

\begin{equation}
\text{Diver}(v_j) = \sum_{i=1}^{k} w_i (v_{ij} - \overline{v_{j}})^2.
\end{equation}
Accordingly, we have the equation:

\begin{equation}
\underbrace{\sum_{i=1}^{k}w_iv_{ij}^2}_{\overline{V^2}} = 
\underbrace{\left (\sum_{i=1}^{k}w_iv_{ij}\right )^2}_{\overline{V}^2} + 
\underbrace{\sum_{i=1}^{k}w_i(v_{ij}-\overline{v_{j}})^2}_{\text{Diver}(V)}\;,\;
j\in [0,C].
\end{equation}
Consequently, $\overline{V^2}$ consists of two parts: $\overline{V}^2$ represents the overall activation intensity within the channel, while $\text{Diver}(V)$ reflects the diversity of features. This indicates that $\overline{V^2}$ provides a more comprehensive and accurate measure of the internal differences within a channel compared to $\overline{V}$. Based on the inverse relationship between $\overline{V^2}$ and $T$, when the activation intensity within a channel is high and feature diversity is pronounced, a smaller $T$ value leads to a sharper weight distribution, causing the model to preferentially select more important features while disregarding non-critical ones. Conversely, when the activation intensity is low and feature diversity is weak, a larger $T$ value results in a more uniform weight distribution, encouraging the model to assess the importance of each token within the channel more evenly.

While both TA-Attention and dual attention mechanisms focus on the differences between channels, they are fundamentally different in their implementation and efficiency. Dual attention mechanisms typically require additional attention modules for the channel dimension, which, although capable of distinguishing channel features, significantly increase computational costs. In contrast, TA-Attention employs a parameter-free approach that adaptively generates the temperature parameter $T$ for each channel by assessing the activation intensity and feature diversity within the channel, resulting in a relatively minor increase in computational load. Additionally, in dual attention mechanisms, spatial and channel attention are two separate modules, with channel attention directly operates on $V$. On the other hand, TA-Attention indirectly regulates the distribution of attention within channels by adjusting the temperature parameter $T$ of the softmax, thereby influencing feature aggregation.

\subsection{MLP-Transformer Hybrid Architecture} \label{sec-MT-hybrid}
MLPs and attention mechanisms complement each other in point cloud analysis. Although local attention can capture intricate dependencies within a localized point region, thereby enhancing the perception of fine-grained features, its complexity remains higher compared to models of similar scale based on MLP. Moreover, Transformer-based models typically demand substantial amounts of data for optimal performance, posing a challenge due to the lack of large-scale datasets for 3D point clouds similar to ImageNet~\cite{ImageNet2009} in the image domain. This scarcity may hinder the training effectiveness of Transformer-based models. Conversely, MLP-based models employ a shared MLP to independently extract features from each point within the perception field, avoiding the modeling of complex relationships between point pairs. Typically, feature aggregation in MLP-based models involves hard-coded pooling operations, rendering the model relatively simple, with low computational complexity and faster convergence. However, these models may struggle to capture local details compared to attention mechanisms.

Given the above analysis, this paper proposes an MLP-Transformer hybrid architecture that combines the complementary strengths of both approaches. In Fig. \ref{fig-Component-Net}, the input to the MLP-Transformer hybrid module undergoes transformation by $\text{FC}_1$ before entering the two parallel MLP and TA-Attention branches. This setup facilitates the extraction of neighborhood features. The MLP branch quickly captures the overall shape information of the current neighborhood through pooling operations, while the local attention branch intricately models the points within the neighborhood and capture fine-grained information. The outputs of these two branches are concatenated and transformed by $\text{FC}_2$ for full integration, serving as the module's output.

During training, the MLP-Transformer hybrid architecture offers a significant advantage over single MLP or Transformer models in terms of improved model convergence speed. At the training stage, the MLP branch swiftly captures local features, guiding the model towards a local optimum and laying a solid information foundation for parameter optimization in the attention branch. Simultaneously, the attention branch enhances the model's representational capacity by extracting fine-grained features, thereby boosting model performance and expediting convergence. Ablation studies validate the advantages of the MLP-Transformer hybrid architecture.

\subsection{Classification Head for Fusing Point and Shape Information} \label{sec-SPF-Head}

This paper devised a novel classification head for the classification model that integrates features at two levels of granularity: point and shape. Termed as the \textit{Shape-Point-Fusion (SPF) Classification Head}, this new classification head operates as follows: assuming the encoder produces $N$ point features $\{x_i\}_{i=1}^{N}$, these $N$ point features are pooled to derive a shape feature $f_{\text{pool}}$. The pooled feature and the $N$ point features are concatenated to form a feature tensor of length $(N+1)$, which is then input into the classification head. The classification head concurrently processes these $(N+1)$ features, yielding $(N+1)$ logits. The $N$ logits corresponding to the point features are averaged to obtain the point-logit, which is then combined with the shape-logit corresponding to the pooled feature to constitute the output of the classification head (refer to Fig. \ref{fig-Component-Net}).

This design enables both point-logit and shape-logit to share the same classification head, with point features directly participating in the classification process. Consequently, during training, backpropagation enforces alignment between the features of individual points and the overall pooled feature, connecting local point features with global context information, thereby allowing local features to be influenced by the overall shape information. This design not only preserves the point features' capability to capture local fine-grained information but also enhances each point's perception of the overall shape in the point cloud.

\subsection{Network Configurations}
Under the PyTorch framework and utilizing an NVIDIA Titan Xp Graphics Processing Unit (GPU), this study developed network architectures focusing on the MT-Block for three common point cloud analysis tasks, with the following specific configurations:

For the point cloud classification model, we employed a three-stage pyramid structure with the MT-Block as the core unit. Each stage contains one block, with downsampling ratios set to 1, 2, and 2, respectively. The neighborhood size increases sequentially to 8, 12, and 16, while the number of output channels scales up from 64 to 256. The training process consists of 90 epochs, where each set of 30 epochs forms an annealing cycle, and the learning rate is annealed from $1.0 \times 10^{-3}$ to $1.0 \times 10^{-5}$.

The part segmentation model adopts the U-Net architecture, with the encoder part featuring a three-stage pyramid structure and incorporating one MT-Block in each stage. Downsampling ratios are set to 1, 4, and 4, while neighborhood sizes range from 16 to 32, and the output channel numbers increase from 128 to 512. The training strategy encompasses 50 epochs with a batch size of 16, employing a non-periodic cosine learning rate strategy, initialized at $1.0 \times 10^{-2}$ and reducing to $1.0 \times 10^{-5}$.

The semantic segmentation model, also based on the U-Net architecture, features an encoder with a five-stage pyramid structure. The MT-Block is deployed in quantities of (1, 2, 3, 2, 2) across the stages, with downsampling ratios and neighborhood sizes set to (1, 4, 4, 4, 4) and (32, 32, 32, 32, 32), respectively. The output channel numbers expand from 32 to 512. The training process employs a non-periodic cosine learning rate, initialized at $1.0 \times 10^{-2}$ and eventually reduced to $1.0 \times 10^{-5}$.

\section{Experiments}
This section offers a thorough assessment of the PointMT model's performance in classification, part segmentation, and semantic segmentation tasks, utilizing widely recognized benchmarks such as ModelNet40, ShapeNet, and S3DIS (Stanford Large-Scale 3D Indoor Spaces). Furthermore, a series of ablation experiments are carried out to validate the effectiveness of the model's design. In the \textbf{supplementary materials}, we provide statistical analysis related to the SPF classification header, as well as some important visualization results pertaining to PointMT.

\subsection{Classification} 
The ModelNet40 dataset consists of 12,311 CAD models spanning 40 categories, utilized for classification, with 9,843 samples for training and 2,468 for testing. Following the PointNet approach \cite{PointNet2017}, we uniformly sample 1,024 points from the surface of each model. Evaluation metrics include overall accuracy (OA) and mean accuracy (mAcc), without employing the voting strategy. The overall accuracy of $94.4\%$, as detailed in Tab. \ref{tab-cls}, demonstrates that the PointMT model matches the performance of state-of-the-art models. Notably, PointMT operates solely with the 1,024 point coordinates as input, unlike other Transformer-based models such as PTv1 \cite{point-trans2021} and PTv2 \cite{point-transV2-2022}, which utilize both coordinates and normals. Despite the class imbalance in ModelNet40, PointMT achieves a class mean accuracy of $92.1\%$, surpassing mainstream models. This underscores PointMT's robustness in handling class imbalance and its ability to extract effective features from limited sample sizes, highlighting the efficacy of the MLP-Transformer hybrid architecture.

\begin{table}[ht]
    \centering
    \footnotesize
    \setlength{\tabcolsep}{0.9mm}
    \begin{tabular}{l|cccc}
        \toprule[1pt]
        \textbf{Method} & Input & \#points & mAcc(\%) & OA(\%) \\
        \midrule[0.3pt]
        \midrule[0.3pt]
        PointNet~\cite{PointNet2017}					& xyz           & 1k    & 86.0  & 89.2  \\
        PointNet++(MSG)~\cite{PointNetplus2017}			& xyz + normal  & 5k    & -     & 91.9  \\
        PointCNN~\cite{PointCNN2018}					& xyz           & 1k    & 88.1  & 92.2  \\
        CSANet~\cite{2022CSANet}                        & xyz           & 1k    & 89.9  & 92.8  \\
        DGCNN~\cite{DGCNN2019}					        & xyz           & 1k    & 90.2  & 92.9  \\
        RS-CNN~\cite{RSCNN2019} w/o vot.                & xyz           & 1k    & -     & 92.9  \\
        KPConv~\cite{KPConv2019}					    & xyz           & 6.8k  & -     & 92.9  \\
        PointGL~\cite{PointGL2024}                      & xyz           & 1k    & 90.4  & 93.0  \\
        PCT~\cite{PCT2021} w/o vot.                     & xyz           & 1k    & -     & 93.2  \\
        PointNext~\cite{PointNext2022}                  & xyz           & 1k    & 90.8  & 93.2  \\
        PointGT~\cite{PointGT2024}                      & xyz           & 1k    & 90.9  & 93.2  \\
        AdaptConv~\cite{Adaptive2021}			        & xyz           & 1k    & 90.7  & 93.4  \\
        PTv1~\cite{point-trans2021}                     & xyz + normal  & 1k    & 90.6  & 93.7  \\
        CurveNet~\cite{CurveNet2021} w/o vot.           & xyz           & 1k    & -     & 93.8  \\
        Auto-Points(w/o voting)~\cite{AutoPoints2024}   & xyz           & 1k    & 91.0  & 93.9  \\
        PTv2~\cite{point-transV2-2022}                  & xyz + normal  & 1k    & 91.6  & 94.2  \\
        MPCT~\cite{MPCT2024}                            & xyz           & 1k    & 92.0  & 94.2  \\
        \midrule[0.3pt]
        PointMT (Ours)                                  & xyz           & 1k    & 92.1  & 94.4  \\
        \bottomrule[1pt]
    \end{tabular}
    \vspace{5pt}
    \caption{Classification results for the ModelNet40 dataset.}
    \label{tab-cls}
\end{table}

\subsection{Part Segmentation}
For part segmentation, we utilize the ShapeNet dataset, comprising 16,880 samples from 16 categories. Each category contains between 2 to 6 parts, totaling 50 distinct parts. Following PointNet\cite{PointNet2017}, we use 2,048 points as input without normals. Evaluation metrics include Intersection over Union (IoU) and mean IoU (mIoU). Experimental results in Tab. \ref{tab-partseg} indicate that PointMT's accuracy is comparable to other models in part segmentation. In comparison with DTNet~\cite{2023DTNet}, which is based on dual attention, this study demonstrates the efficacy of the TA-Attention and hybrid architecture presented herein for feature extraction.

\begin{table}[ht]
    \centering
    \footnotesize
    \setlength{\tabcolsep}{3.5mm}
    \begin{tabular}{l|cc} 
        \toprule[1pt]
        \textbf{Method} & mIoU(\%) & IoU(\%)  \\
        \midrule[0.3pt]
        PointNet~\cite{PointNet2017}        & 80.4  & 83.7 \\
        PointNet++~\cite{PointNetplus2017}  & 81.9  & 85.1 \\
        DGCNN~\cite{DGCNN2019}              & 82.3  & 85.2 \\
        PointGL~\cite{PointGL2024}          & 83.8  & 85.6 \\       
        AdaptConv~\cite{Adaptive2021}       & 83.4  & 86.4 \\
        KPConv~\cite{KPConv2019}            & 85.1  & 86.4 \\
        PTv1~\cite{point-trans2021}         & 83.7  & 86.6 \\    
        PointGT~\cite{PointGT2024}          & 84.7  & 86.7 \\
        MPCT~\cite{MPCT2024}                & -     & 86.7 \\
        Auto-Points~\cite{AutoPoints2024}   & 85.0  & 87.1 \\
        \midrule[0.3pt]
        PointMT (Ours)                      & 83.6  & 85.9 \\
        \bottomrule[1pt]
    \end{tabular}
    \vspace{5pt}
    \caption{Part segmentation results for the ShapeNetPart dataset.}
    \label{tab-partseg}
\end{table}

\subsection{Semantic Segmentation}
The S3DIS dataset encompasses extensive indoor scenes from six areas within three buildings, comprising a total of 271 rooms. Each point contains $xyz$ coordinates and RGB color information, along with a semantic label. With 13 semantic labels such as floors, tables, and ceilings, we designate Area-5 as the test set, utilizing samples from other areas for training. Evaluation metrics encompass class-wise Intersection over Union (mIoU), mean accuracy (mAcc), and overall accuracy (OA). The results, as depicted in Tab. \ref{tab-s3dis}, illustrate that PointMT, characterized by its lightweight architecture, outperforms other conventional methods, notably surpassing the largest variant of the advanced model PointNext~\cite{PointNext2022}, identified as PointNext-XL. With a parameter count of 11.4 M, PointMT exhibits superior parameter efficiency compared to other state-of-the-art models. Its reduced spatial complexity enhances performance in practical scene-based semantic segmentation tasks, thereby improving deployability.

\begin{table}[ht]
    \centering
    \footnotesize
    \setlength{\tabcolsep}{2.0mm}
    \begin{tabular}{l|c|ccc}
        \toprule[1pt]
        \textbf{Method}                      & Params   & mIoU  & OA    & mAcc  \\
        \midrule[0.3pt]
        PointNet~\cite{PointNet2017}         & 3.6 M    & 41.1  & –     & 49.0  \\
        SegCloud~\cite{SEGCloud2017}         &  -       & 48.9  & –     & 57.4  \\
        PointCNN~\cite{PointCNN2018}         & 0.6 M    & 57.3  & 85.9  & 63.9  \\
        PCCN~\cite{PCNN2018}                 & -        & 58.3  & –     & 67.0  \\
        PointWeb~\cite{PointWeb2019}         & -        & 60.3  & 87.0  & 66.6  \\
        PointASNL~\cite{PointASNL2020}       & -        & 62.6  & 87.7  & 68.5  \\
        PointGL~\cite{PointGL2024}           & 3.5 M    & 65.6  & 88.6  & 71.9  \\
        KPConv~\cite{KPConv2019}             & 15.0 M   & 67.1  & –     & 72.8  \\
        AdaptConv~\cite{Adaptive2021}        & -        & 67.9  & 90.0  & 73.2  \\ 
        MPCT~\cite{MPCT2024}                 & -        & 68.6  & 89.3  & 74.6  \\
        Auto-Points~\cite{AutoPoints2024}    & 23.4 M   & 69.1  & 90.2  & -     \\
        PTv1~\cite{point-trans2021}          & -        & 70.4  & 90.8  & 76.5  \\
        PointNext-XL~\cite{PointNext2022}    & 41.6 M   & 70.5  & 90.6  & 76.8  \\
        PTv2 \cite{point-transV2-2022}       & -        & 71.6  & 91.1  & 77.9  \\       
        \midrule[0.3pt]
        PointMT (Ours)                       & 11.4 M    & 71.0  & 90.9  & 78.6 \\ 
        \bottomrule[1pt]
    \end{tabular}
    \vspace{5pt}
    \caption{Semantic segmentation results for the S3DIS (Area-5) dataset (\%).}
    \label{tab-s3dis}
\end{table}

\subsection{Architecture Design}
This paper presents an experimental evaluation of the PointMT model, which integrates a temperature adaptation attention mechanism with an MLP-Transformer hybrid module in its foundational architecture. Furthermore, for classification tasks, the model incorporates a classification head that merges global pooling features with point features. A series of ablation experiments are conducted to validate the model's design effectiveness, with results detailed in Tab. \ref{tab-pointmt-ablation-design}.

\begin{table}[ht]
    \centering
    \small
    \newcommand{\tabincell}[2]{\begin{tabular}{@{}#1@{}}#2\end{tabular}}
    \setlength{\tabcolsep}{0.6mm}
    \begin{tabular}{l|cc|cc|cc|c}
        \toprule[1pt]
        Model    &\tabincell{c}{Attn\\(Linear)}         &\tabincell{c}{Attn\\(Linear\\+TA)}    
                 &\tabincell{c}{Block\\(TA-Attn)}       &\tabincell{c}{Block\\(Hybrid)}  
                 &\tabincell{c}{Cls.\\Head\\(Trad.)} &\tabincell{c}{Cls.\\Head\\(SPF)}
                 & \tabincell{c}{Acc.\\(\%)}            \\
        \midrule[0.3pt]
        Model-A         &\ding{52}   &            &            &\ding{52}    &             &\ding{52}   & 93.4   \\
        Model-B         &            &\ding{52}   &\ding{52}   &             &             &\ding{52}   & 92.5   \\
        Model-C         &            &\ding{52}   &            &\ding{52}    &\ding{52}    &            & 93.8   \\
        \midrule[0.3pt]
        \tabincell{c}{PointMT\\(Ours)}
                    &            &\ding{52}   &            &\ding{52}    &             &\ding{52}   & 94.4   \\
        \bottomrule[1pt]
    \end{tabular}
    \vspace{5pt}
    \caption{Ablation study of the architecture design. This table illustrates the impact of various components on model performance. ``Attn(Linear)" refers to the local attention mechanism with linear complexity introduced in Sec. \ref{sec-linear-attn};``Attn(Linear+TA)" denotes the temperature adaptation mechanism integrated into the linear attention; ``Block(TA-Attn)" indicates a block solely incorporating TA-Attention; ``Block(Hybrid)" signifies a block integrating the MLP-Transformer hybrid structure; for the classification head, ``Trad." represents the conventional classification head based on pooling operations, while ``SPF" denotes a novel classification head that takes the ensemble of shape and point features as input.}
    \label{tab-pointmt-ablation-design}
\end{table}

The integration of the TA-Attention, the MT-Block, and the SPF classification head design within the PointMT model demonstrates a significant enhancement in point cloud analysis tasks' performance. Ablation experiments confirm the model's design advantages, highlighting its potential for practical applications in AI-driven 3D data processing.

\subsection{Complexity Analysis}
In the pursuit of improving model performance, PointMT emphasizes the importance of reducing model complexity. The introduction of local linear attention significantly decreases the model's time complexity. The parameter-free temperature adaptation mechanism adjusts the weight distribution in each channel via $T$ without increasing spatial complexity. Incorporating an additional MLP branch in the MLP-Transformer hybrid module design has significantly enhanced performance. Although the MLP branch adds to the computational load, its complexity remains considerably lower compared to the parallel attention branches. As demonstrated by Tab. \ref{tab-complexity}, in comparison to other Transformer-based methods~\cite{point-trans2021, point-transV2-2022}, PointMT achieves superior accuracy, striking an optimal balance between performance and complexity.

\begin{table}[ht]
    \centering
    \newcommand{\tabincell}[2]{\begin{tabular}{@{}#1@{}}#2\end{tabular}}
    \footnotesize
    \setlength{\tabcolsep}{2.0mm}
    \begin{tabular}{l|cccc}
        \toprule[1pt]
        \textbf{Method}     & \tabincell{c}{Params.\\(M)}   & \tabincell{c}{FLOPs\\(G)}
                            & \tabincell{c}{mAcc\\(\%)}     & \tabincell{c}{OA\\(\%)} \\  
        \midrule[0.3pt]
        PointNet~\cite{PointNet2017}					& 3.5           & 0.9   & 86.2  & 89.2 \\
        PointNet++(MSG)~\cite{PointNetplus2017}			& 1.7           & 4.1   & -     & 91.9 \\
        PointCNN~\cite{PointCNN2018}					& 0.6           & -     & 88.1  & 92.2 \\
        DGCNN~\cite{DGCNN2019}					        & 1.8           & 4.8   & 90.2  & 92.9 \\
        PointGL~\cite{PointGL2024}                      & 4.2           & 0.6   & 90.4  & 93.0 \\
        PointNext-S~\cite{PointNext2022}                & 4.5           & 6.5   & 90.9  & 93.7 \\
        PTv1~\cite{point-trans2021}                     & 11.4          & -     & -     & 93.7 \\
        Auto-Points(w/o voting)~\cite{AutoPoints2024}   & 5.1           & 6.0   & 91.0  & 93.9 \\
        PTv2~\cite{point-transV2-2022}                  & 12.8          & -     & -     & 94.2 \\
        PointWavelet-L~\cite{PointWavelet2023}          & 58.4          & 39.2  & 91.1  & 94.3 \\
        \midrule[0.3pt]
        PointMT (Ours)                                  & 2.4           & 2.7   & 92.1  & 94.4 \\
        \bottomrule[1pt]
    \end{tabular}
    \vspace{5pt}
    \caption{Complexity Analysis of PointMT on the ModelNet40 Classification Task. (M: $10^{6}$, G: $10^{9}$).}
    \label{tab-complexity}
\end{table}

\subsection{Effect of the MLP-Transformer Architecture}
The PointMT model proposed in this study demonstrates notable characteristics of rapid convergence, indicating its capability to achieve superior performance within fewer training iterations, thus reducing computational resource consumption. This advantage primarily arises from the hybrid architecture of MLP-Transformer. To compare convergence rates among different model structures, we conducted experiments using an annealing learning rate over cycles of 30 epochs, comparing accuracy and loss values under three conditions: a model with only the MLP branch, another with only the Attention branch, and a hybrid architecture of MLP-Transformer.

\begin{figure}[ht]
    \centering
    \includegraphics[width=0.7\linewidth]{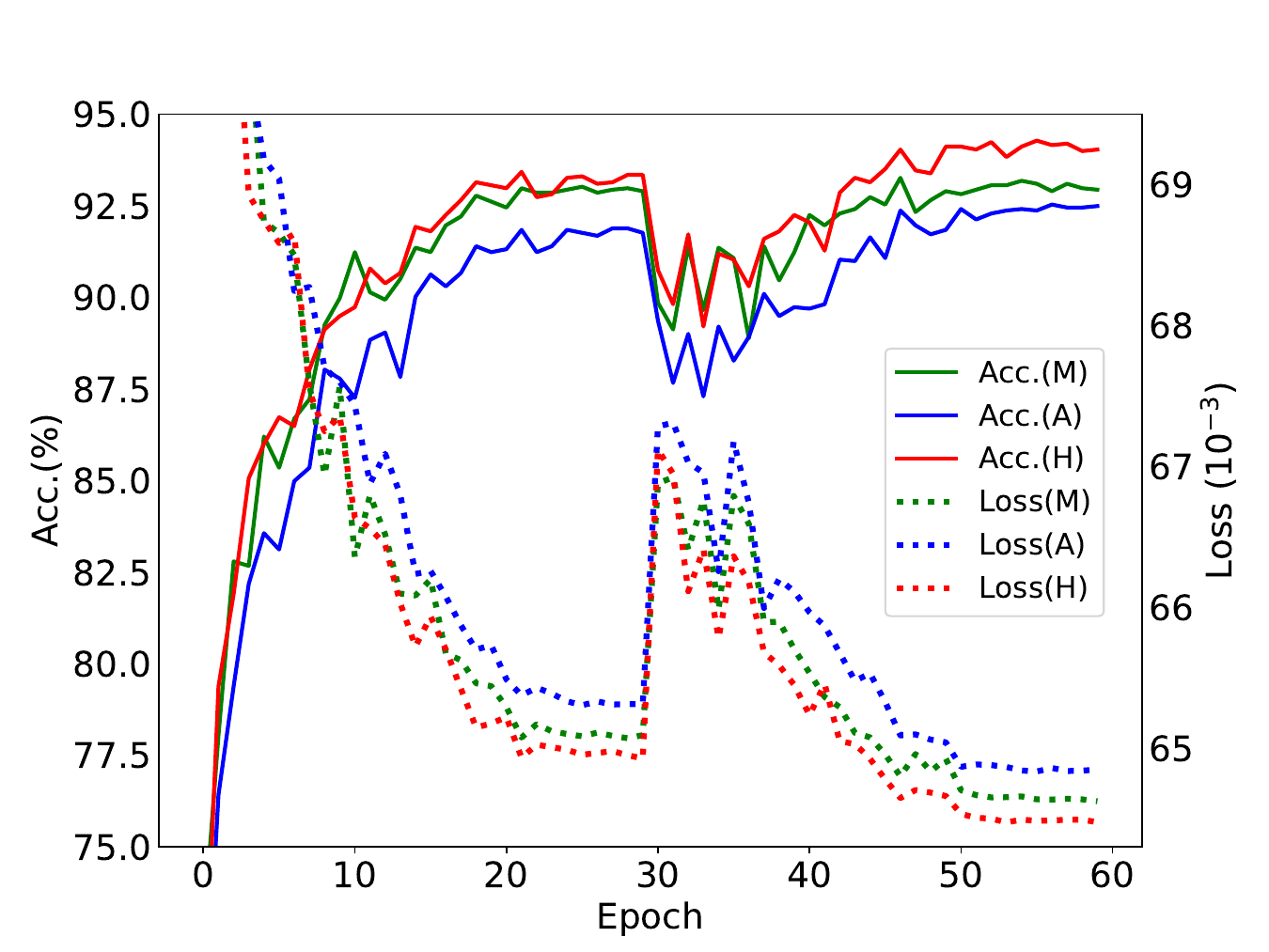}
    \caption{Comparison of convergence rates among different model architectures (M: MLP, A: Attention, H: MLP-Transformer hybrid). The hybrid architecture achieves notable accuracies of $93.4\%$ and $94.2\%$ after 30 and 60 epochs, respectively, indicating rapid convergence.}
    \label{fig-acc-loss}
\end{figure}

Fig. \ref{fig-acc-loss} presents the experimental results, comparing data from the first two annealing cycles. Analysis of the figure data reveals that while the MLP-only model exhibits relatively fast convergence, its overall performance is limited due to the insufficient capability of MLP in capturing local fine-grained features. Conversely, the Transformer-based model demonstrates slower convergence, with limited accuracy improvement within a finite training set scale. However, the model employing the MLP-Transformer hybrid module can leverage the MLP network to quickly approach the local optimal solution during the initial training stages, and later utilize the attention mechanism to further refine fine-grained features, effectively enhancing model performance. As depicted in the first and second cycles of the figure, the model already achieved accuracies of $93.4\%$ and $94.2\%$ within an exceptionally brief period, respectively, surpassing many previous advanced models. The trend in the change of loss value indirectly confirms this conclusion, suggesting that the MLP-Transformer hybrid architecture effectively amalgamates the advantages of MLP and Transformer, not only enhancing model performance but also significantly increasing convergence rates.

\subsection{Neighborhood Size}
The size of the neighborhood is a hyperparameter that plays a critical role in performance. In this paper, our classification model employs a pyramidal encoder composed of three stages. Unlike other works, the neighborhood size $k$ corresponding to these three stages is set to 8, 12, and 16, respectively, following a linearly increasing trend. The results presented in Tab. \ref{PointMT-k} demonstrate the outcomes when the $k$ values for the three stages are uniformly set to a fixed value. It is evident that gradually increasing the neighborhood size significantly improves the accuracy of classification.

\begin{table}[ht]
    \centering
    \small
    \setlength{\tabcolsep}{2.0mm}
    \begin{tabular}{c|cccc}
        \toprule[1pt]
        $k$              & [8, 8, 8]  	& [12, 12, 12]  	& [16, 16, 16] 	& [8, 12, 16] \\
        \midrule[0.3pt]
         Acc. (\%) 	    & 93.4 	& 93.5  & 93.8  & 94.4 \\ 
        \bottomrule[1pt]
    \end{tabular}
    \vspace{5pt}
    \caption{ModelNet40 classification results with varying neighborhood combinations (three sizes per group corresponding to the encoder's three stages).}
    \label{PointMT-k}
\end{table}

The advantage of this design strategy lies in the fact that smaller $k$ values at the shallow layers of the pyramid effectively capture local details of the point cloud, while larger $k$ values at deeper layers facilitate the integration of global features. This hierarchical feature extraction strategy allows the model to capture rich spatial information at different levels, compensating for the inability of local attention mechanisms to directly perceive global information, thereby achieving more refined recognition in point cloud classification tasks. 

Additionally, such a design effectively balances computational efficiency and performance. As analyzed in Sec. \ref{sec-linear-attn}, this paper adopts an attention mechanism with a complexity of $\mathcal{O}(Nk)$. At the shallow layers of the pyramidal structure, where the number of points $N$ is larger, a smaller $k$ value reduces the computational load introduced by the attention mechanism. As the network deepens and the point cloud undergoes downsampling, the number of points $N$ gradually decreases, and increasing $k$ at this stage does not significantly increase computational costs. Therefore, this design maintains the model's efficient operation while ensuring an effective expansion of the perception field.

Moreover, the encoder at the shallow layers primarily extracts low-level geometric features, which are sensitive to noise and outliers. Since point cloud data is relatively dense and features are simpler at this stage, adopting a smaller $k$ value allows the KNN algorithm to more accurately identify the nearest neighborhood points to the target point, avoiding excessive influence from noise. At deeper layers, a larger $k$ value smooths out these local disturbances, capturing more contextual information in sparse data and enhancing the model's robustness to noise.

\subsection{Visualization Analysis of Encoder Features and Logit Outputs}

\begin{figure}[ht]
    \centering
    \includegraphics[width=0.95\linewidth]{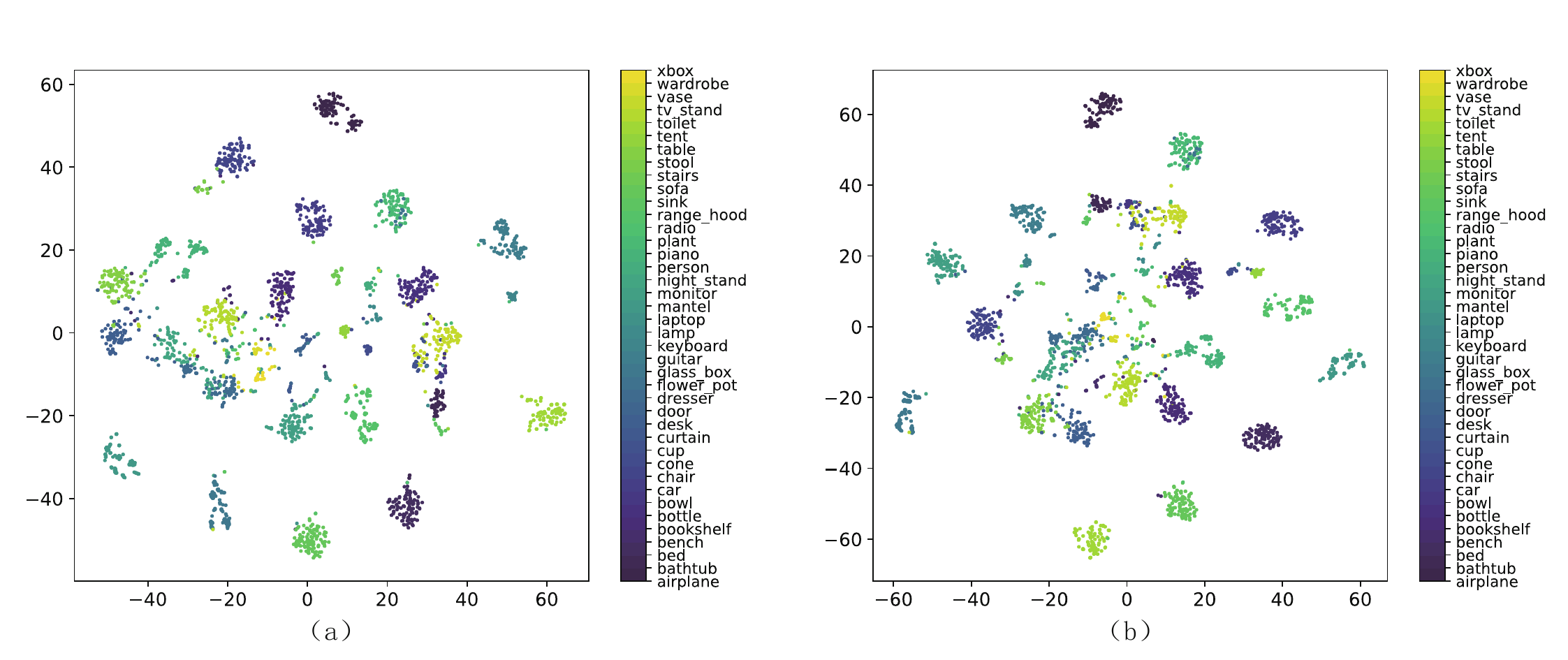}
    \caption{t-SNE visualization of encoder features using (a) the conventional classification head and (b) the SPF classification head.}
    \label{PointMT-fig-t-SNE-feature}
\end{figure}

\begin{figure}[ht]
    \centering
    \includegraphics[width=0.95\linewidth]{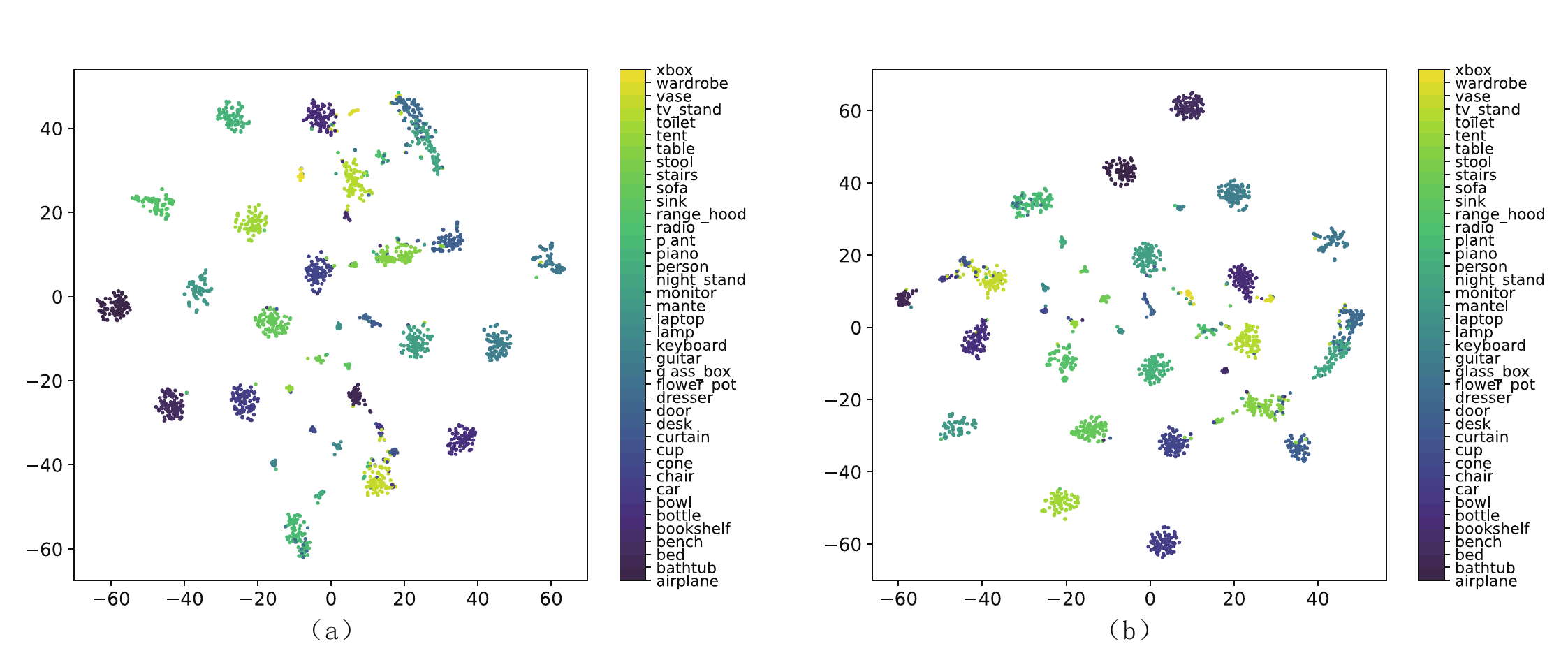}
    \caption{t-SNE visualization of logit outputs using (a) the conventional classification head and (b) the SPF classification head.}
    \label{PointMT-fig-t-SNE-logit}
\end{figure}

To verify the effectiveness of the SPF classification head, we conducted ablation experiments comparing it with a traditional classification head, using t-SNE (t-Distributed Stochastic Neighbor Embedding) for visual analysis of both encoded features and logit outputs. The results, shown in Fig. \ref{PointMT-fig-t-SNE-feature} and Fig. \ref{PointMT-fig-t-SNE-logit}, are used to explore the differences in feature representation and classification performance between the two models.

As illustrated in Fig. \ref{PointMT-fig-t-SNE-feature}, the t-SNE visualizations of the encoded features for the two models do not exhibit significant differences. This outcome can be attributed to the fact that these encoded features primarily represent the shape features obtained through pooling, without incorporating the point-level features that are fed into the SPF classification head and play a direct role in classification. Although quantitative experiments (Sec. \ref{PointMT-sec-SPF}) conducted later in this study demonstrated that the SPF classification head leads to notable differences in point features, these differences are not sufficiently pronounced to cause significant variation in the pooled shape features. Therefore, the t-SNE visualization of encoded features alone does not fully capture the impact of the classification heads on overall model performance.

The more meaningful comparison comes from the t-SNE visualization of the logit outputs, which directly reflect the classification outcomes, as shown in Fig. \ref{PointMT-fig-t-SNE-logit}. In this figure, clear differences emerge between the two models. In Fig. \ref{PointMT-fig-t-SNE-logit}(b), which corresponds to the SPF classification head, the clusters are more compact, and some exhibit a nearly circular distribution, indicative of higher classification confidence. Moreover, the clusters are more distinct, with clearer separations between them and fewer stray points, signifying more precise decision boundaries and improved generalization ability. These results demonstrate that the integration of shape and point features in the SPF classification head leads to enhanced classification performance, with greater confidence and superior generalization. Thus, the SPF classification head proves to be more effective in achieving robust and confident classification outcomes compared to the traditional approach.

\subsection{Visual Analysis of the SPF Classification Head}
This section delves into the performance advantages of the Shape-Point Fusion (SPF) classification head for point cloud classification through ablation studies. Two models are compared: Model A utilizes a traditional classification head where pooled shape features serve as inputs, serving as the baseline experiment; Model B employs the SPF classification head, concatenating $N$ point features with one pooled feature to form ($N+1$) features for the classification head input. Both scenarios involve saving corresponding pre-trained models, and the test set is used for inference. \textbf{During inference, both the pre-trained Model A and B} employ the concatenation of point features and pooled features as the input of its classification head, saving the shape-logit corresponding to the pooled features and the $N$ point-logits corresponding to the point features. Based on these saved logits, three experiments are designed to validate the performance gains brought by the SPF classification head to classification tasks.

\subsubsection{Statistical Analysis of KL Divergence}
In this study, we utilized the softmax function on both the shape-logit and the $N$ point-logits to derive probabilistic predictions. Model B incorporating the Shape-Point Fusion (SPF) classification head integrates point features and pooled features during the training phase, with these features merging at a shared classification head. Conversely, Model A trained with traditional classification heads, while capable of temporarily transplanting the SPF classification head during inference to obtain the shape-logit and $N$ point-logits, do not directly involve point features in the predictive process during training.

\begin{figure}[htbp]
    \centering
    \includegraphics[width=0.8\linewidth]{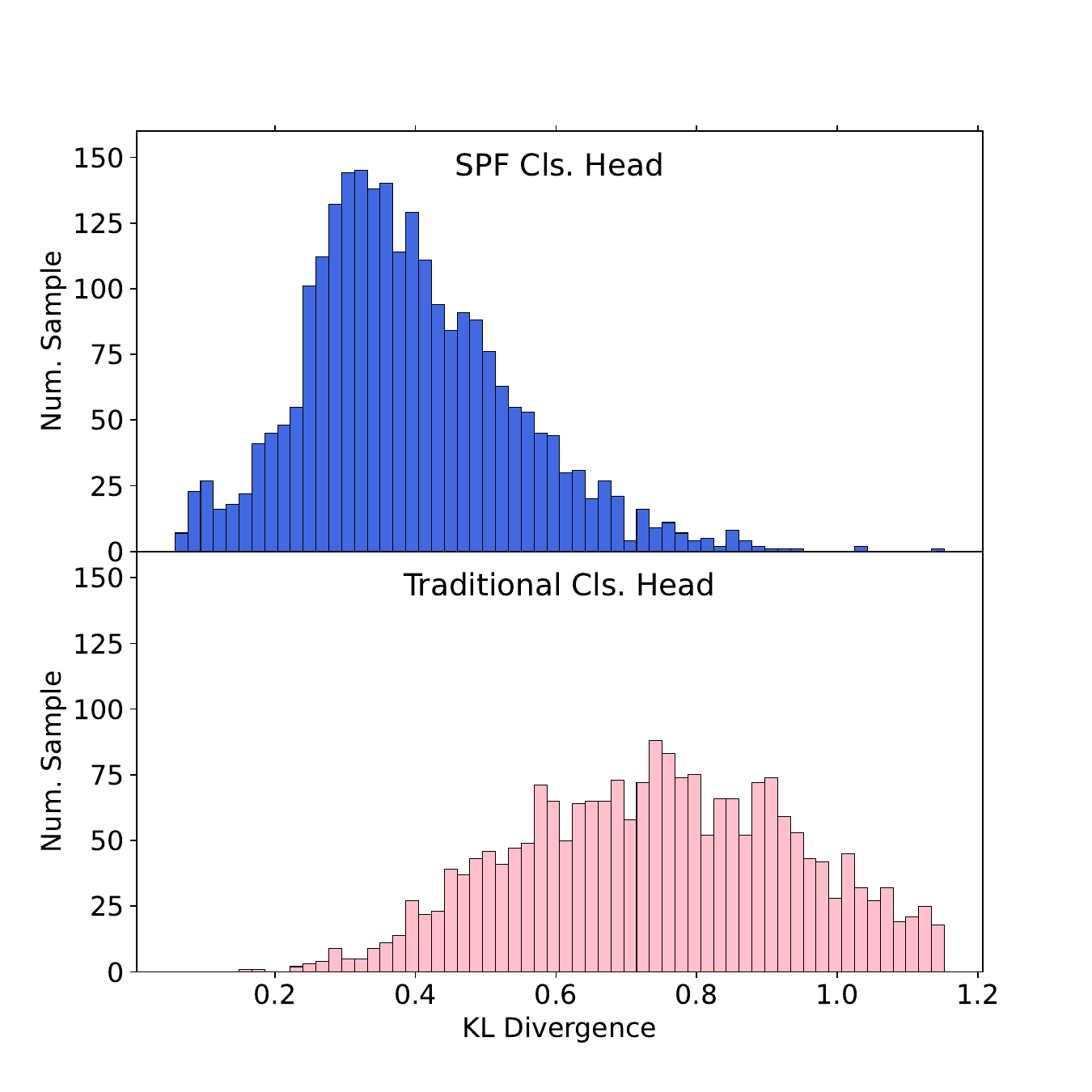}
    \caption{Distribution of sample count versus KL values for models with SPF and traditional classification head. We adopt the mean KL divergence between the shape-logit and the $N$ point-logits of each sample in the test set as the criterion for model performance evaluation. Subsequently, we aggregated the KL divergence values across the entire test set and constructed a statistical histogram to visually represent the distribution of performance.}
    \label{fig-PointMT-KL-statistics}
\end{figure}

For each sample in the test set, we computed the average Kullback-Leibler (KL) divergence between the predicted probabilities of $N$ point features and one shape feature, establishing this as the KL metric for each sample. Figure \ref{fig-PointMT-KL-statistics} presents the distribution of sample counts against the KL values. The model (Model B) trained with the SPF classification head exhibits a prominent leftward peak, indicating lower KL divergences between the point-logit and shape-logit predictions for each individual sample. Further computational results reveal that the average KL divergences in the test set for the two configurations are 0.39 and 0.82, respectively.

The analysis indicates that when employing the SPF classification head, there is an increased alignment in the predictive outcomes of point features and pooled features, signifying improved feature consistency. This consistency enables point features to implicitly assimilate global information, thereby enhancing their discriminative capabilities.

\subsubsection{Statistical Analysis of Discriminability of Point Features}
In the PointMT model, each point can only access limited global information due to its local perceptual field, it becomes necessary to aggregate the $N$ point-logits before merging them with the shape-logit to generate the final output. This preliminary step serves to address the inadequacy of information among the point-logits. 

\begin{figure}[htbp]
    \centering
    \includegraphics[width=0.8\linewidth]{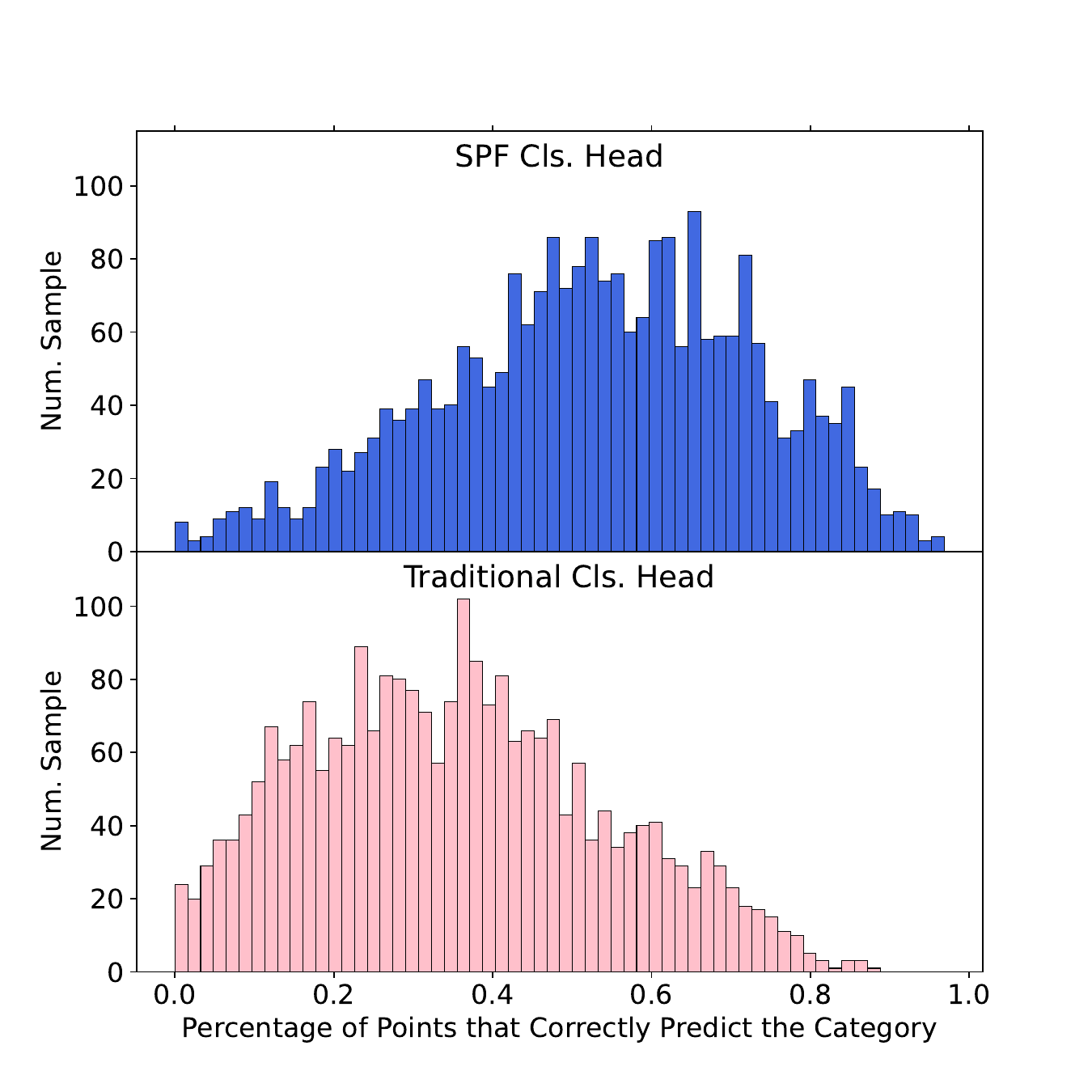}
    \caption{Distribution of sample count versus the percentage of points accurately predicting the category for models employing the SPF classification head and traditional classification head. We analyzed each sample in the test set to count the number of points that accurately predict the shape category, using the resulting percentage as the criterion. After conducting this statistical analysis, we compiled the data for the entire test set and plotted a histogram to visually represent the distribution.}
    \label{fig-PointMT-correct_point-statistics}
\end{figure}

In this section, we compute the percentage of point-logits accurately predicting the shape for each test sample and subsequently analyze the statistical relationship between the sample count and this percentage, as illustrated in Fig. \ref{fig-PointMT-correct_point-statistics}. The figure depicts a notable rise in the proportion of points accurately predicting the category for models trained with the SPF classification head, indicated by a shift in the peak towards the right. Additionally, we calculates the average percentage of such points across all samples for the two models, which are $55.0\%$ and $36.6\%$, respectively. This improvement is attributed to the SPF classification head's capacity to merge point features with global features, focusing not only on local intricacies but also encompassing contextual information regarding the overall shape. This strategy empowers individual points to acquire features with heightened category specificity.

\subsubsection{Visualization of Point Feature Discriminability}
\begin{figure}[htbp]
    \centering
    \includegraphics[width=0.7\linewidth]{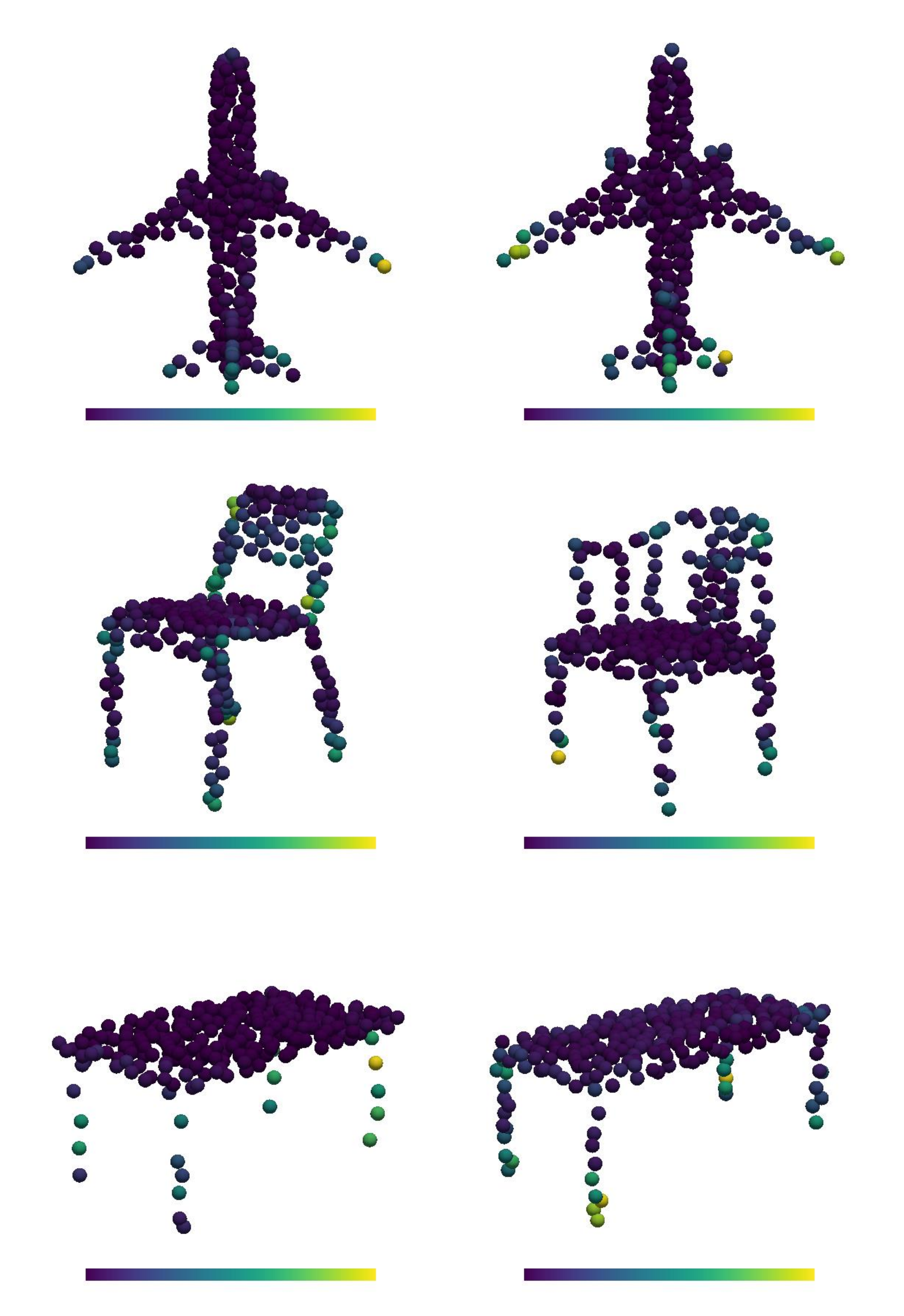}
    \caption{Visualization of the KL divergence of point-logits. The figure displays four rows of data, with each row representing two samples from the same category. The color of each point corresponds to the KL divergence between the point's predicted outcome and the shape's one-hot label.}
    \label{fig-PointMT-vis-KL}
\end{figure}

We assess the discriminative capability of each point within the each sample by calculating the KL divergence between these probabilities and the one-hot label. To evaluate this, we selected multiple samples from the test set and visualized the KL divergence within each sample, as illustrated in Fig. \ref{fig-PointMT-vis-KL}. Each row in the visualization juxtaposes two samples from the same class. The findings reveal that critical positions for class determination, such as the fuselage and wings of an airplane, the seating surface of a chair, and the tabletop of a table, often correspond to points with lower KL divergence. This suggests that these points harbor certain global information, empowering them to make confident judgments concerning the object's category.

\subsection{Visual Analysis of TA-Attention}
The PointMT model integrates a temperature adaptation mechanism to enhance feature aggregation. This mechanism adjusts the softmax parameter $T$ for each channel, generating distinct attention weights for each center point within the sample across \(k\) neighboring points and \(C\) channels, resulting in an \(N \times C\) weight matrix. This section illustrates the distribution of the TA-Attention weights through heatmap visualizations.

\begin{figure}[htbp]
    \centering
    \includegraphics[width=0.99\linewidth]{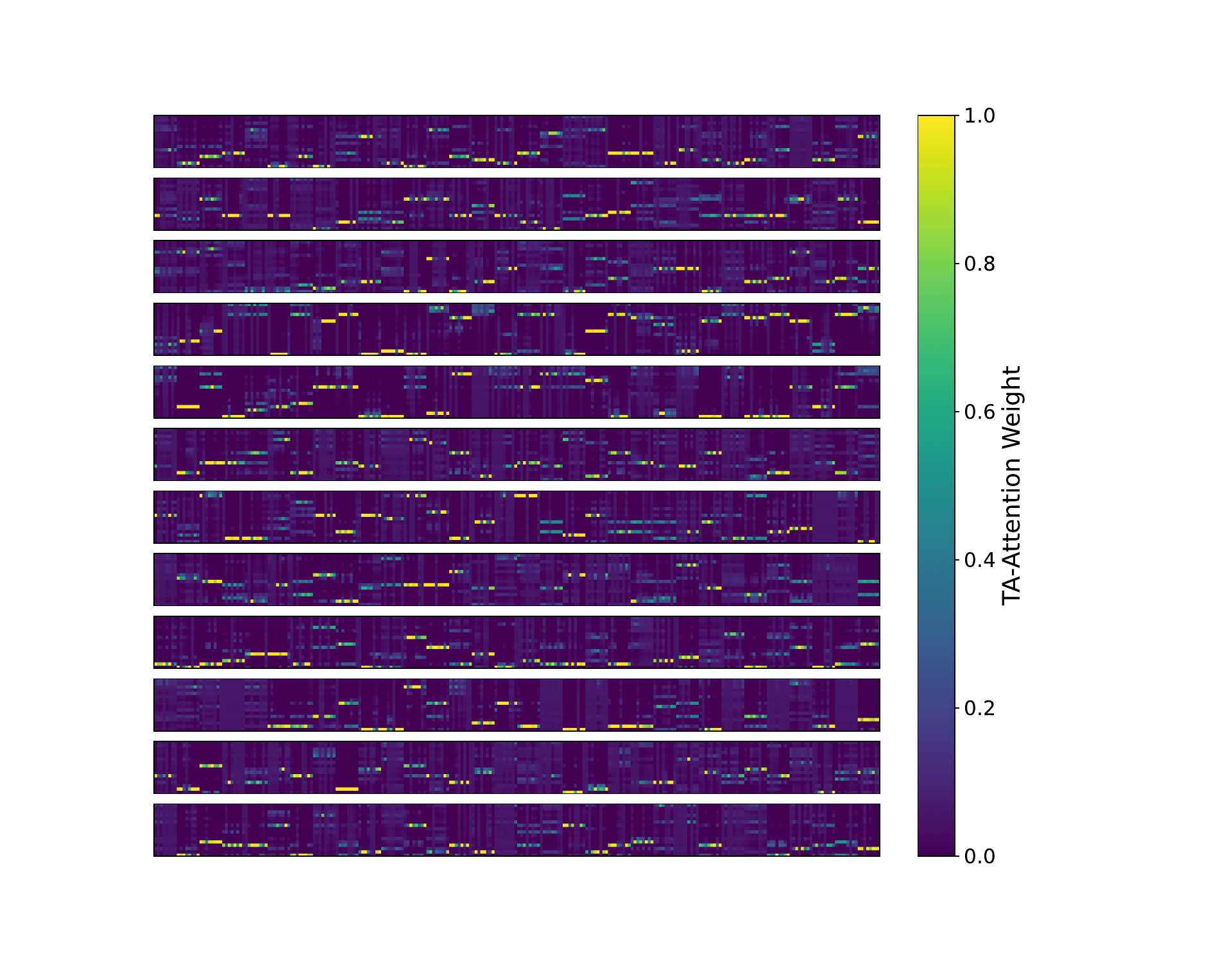}
    \caption{Visualization of TA-attention attention weights. The figure demonstrates the local attention patterns for 12 points within a sample, where each set of weights is depicted by a $16\times256$ matrix, corresponding to the attention distribution across 16 neighborhood points and 256 channels.}
    \label{fig-PointMT-vis-TA-attn}
\end{figure}

To delve deeper into the TA-Attention mechanism, we selected the attention weights of 12 points from the test samples for visual analysis, as illustrated in Fig. \ref{fig-PointMT-vis-TA-attn}. Each point considers 16 neighboring points with 256 feature channels. Upon examining the heatmap, it becomes apparent that the spatial attention weights vary significantly across different channels. This suggests that relying solely on spatial attention is inadequate for capturing the disparities between channels. In certain channels, the attention weight for a specific neighboring point approaches 1, while others have weights close to 0, resembling max pooling behavior in this context. Conversely, there are channels exhibiting a more uniform distribution of attention weights across neighboring points, similar to average pooling behavior.

These visualizations illustrate that the TA-Attention mechanism considers the interrelations of features in both spatial and channel dimensions, dynamically generating weights. This mechanism not only enriches the representation of crucial features but also suppresses irrelevant information. Compared to max pooling and average pooling, the TA-Attention mechanism offers greater adaptability and precision in feature aggregation.

\section{Conclusion}
This study introduces PointMT, a novel architecture for point cloud analysis based on an MLP-Transformer hybrid model, designed to efficiently extract features from point clouds for various multimedia applications. The key contributions of PointMT include a local attention mechanism with linear complexity and the introduction of a temperature adaptation mechanism. This integration allows for precise extraction of fine-grained features without increasing model size. Additionally, to address convergence issues commonly encountered in Transformer models, the MLP-Transformer hybrid architecture significantly enhances convergence speed and performance. For point cloud classification tasks, a classification head is introduced that merges point and shape features, thereby enhancing discriminative capabilities. PointMT achieves a balance between performance and complexity, aligning with state-of-the-art models' performance while maintaining efficiency.

However, this study has certain limitations. Despite the introduction of a pyramid structure to address the lack of global information perception in local attention mechanisms, the PointMT model still lacks direct global perception, which may limit its application in complex multimedia scenarios. Additionally, while the TA-Attention mechanism itself does not increase the model's complexity, the intermediate variables involved in its computation require a certain amount of memory space. Furthermore, the performance of the PointMT model in other point cloud analysis tasks, particularly those specific to multimedia applications, remains to be fully evaluated. Future research should focus on developing attention mechanisms that are more computationally efficient and perceptually capable, and on exploring the potential applications of the PointMT model in a broader range of scenarios, including those prevalent in multimedia processing and related fields.

\bibliographystyle{IEEEtran}

\bibliography{reference}

\newpage

\vfill

\end{document}